\documentclass[letterpaper]{article} 
\usepackage{aaai25}  
\usepackage{times}  
\usepackage{helvet}  
\usepackage{courier}  
\usepackage[hyphens]{url}  
\usepackage{graphicx} 
\usepackage{natbib}  
\usepackage{caption} 
\usepackage{algorithm}
\usepackage{algorithmic}

\usepackage{newfloat}
\usepackage{listings}
\DeclareCaptionStyle{ruled}{labelfont=normalfont,labelsep=colon,strut=off} 
\floatstyle{ruled}
\newfloat{listing}{tb}{lst}{}
\floatname{listing}{Listing}
\usepackage{subcaption}
\usepackage{amsmath}
\usepackage{booktabs}
\usepackage{makecell}
\usepackage{colortbl}
\usepackage{xcolor}
\usepackage{multirow}

\definecolor{mygray}{gray}{.92}

\title{COMM: Concentrated Margin Maximization for Robust Document-Level Relation Extraction}
\author{
    Zhichao Duan\textsuperscript{\rm 1},
    Tengyu Pan\textsuperscript{\rm 1},
    Zhenyu Li\textsuperscript{\rm 1},
    Xiuxing Li\textsuperscript{\rm 2},
    Jianyong Wang\textsuperscript{\rm 1}\thanks{Corresponding author}
}
\affiliations{
    \textsuperscript{\rm 1}Tsinghua University\\
    \textsuperscript{\rm 2}Beijing Institute of Technology\\

}

\usepackage{bibentry}

\begin{document}

\maketitle

\begin{abstract}
Document-level relation extraction (DocRE) is the process of identifying and extracting relations between entities that span multiple sentences within a document. 
Due to its realistic settings, DocRE has garnered increasing research attention in recent years. 
Previous research has mostly focused on developing sophisticated encoding models to better capture the intricate patterns between entity pairs. While these advancements are undoubtedly crucial, an even more foundational challenge lies in the data itself. 
The complexity inherent in DocRE makes the labeling process prone to errors, compounded by the extreme sparsity of positive relation samples, which is driven by both the limited availability of positive instances and the broad diversity of positive relation types.
These factors can lead to biased optimization processes, further complicating the task of accurate relation extraction. 
Recognizing these challenges, we have developed a robust framework called \textit{\textbf{COMM}} to better solve DocRE.
\textit{\textbf{COMM}} operates by initially employing an instance-aware reasoning method to dynamically capture pertinent information of entity pairs within the document and extract relational features. 
Following this, \textit{\textbf{COMM}} takes into account the distribution of relations and the difficulty of samples to dynamically adjust the margins between prediction logits and the decision threshold, a process we call Concentrated Margin Maximization.
In this way, \textit{\textbf{COMM}} not only enhances the extraction of relevant relational features but also boosts DocRE performance by addressing the specific challenges posed by the data. 
Extensive experiments and analysis demonstrate the versatility and effectiveness of \textit{\textbf{COMM}}, especially its robustness when trained on low-quality data (achieves \textgreater 10\% performance gains).
\end{abstract}

%
\section{Introduction}\label{intro}

In recent years, the task of document-level relation extraction (DocRE) has garnered significant research attention due to its realistic and practical settings \cite{duan-etal-2022-just,wang-etal-2023-adaptive,xu2024document,xiao-etal-2024-federated-document}. 
Unlike sentence-level relation extraction \cite{lyu2021relation, zhou2022improved}, DocRE involves identifying and extracting relations between entities that span multiple sentences within a document \cite{yao-etal-2019-docred}. 
This complexity reflects more closely the natural flow of information in real-world documents, such as news articles \cite{yamamoto2017company}, scientific papers \cite{jain-etal-2020-scirex}, and financial reports \cite{oral2020information}, making DocRE a crucial task for various natural language processing (NLP) applications.


\begin{figure}[t!]
    \centering
    \includegraphics[width=\linewidth]{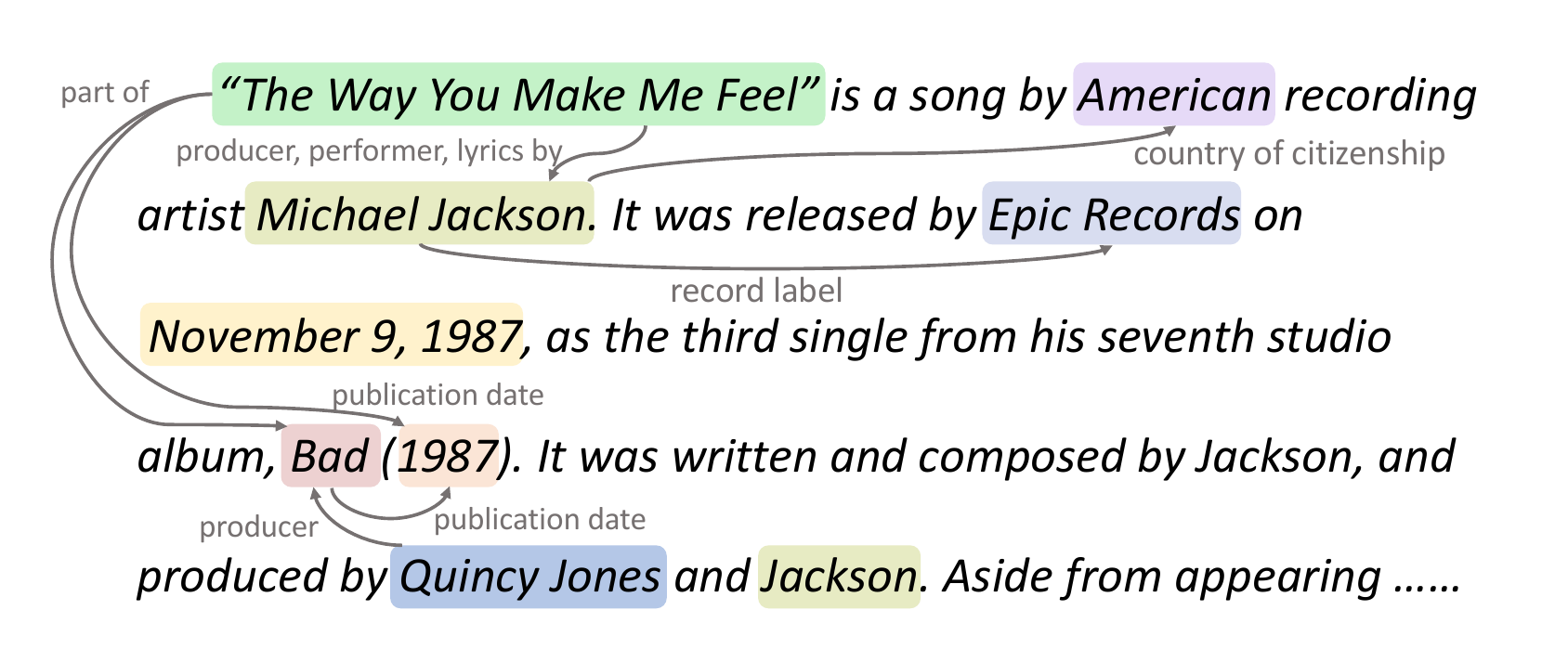}
    \caption{A sample document in Re-DocRED \cite{tan-etal-2022-revisiting}. Different entities are highlighted in distinct colors, with all mentions of the same entity sharing the same color. For brevity, some sentences and relations have been omitted from the illustration.}
    \label{fig:example}
\end{figure}

\begin{figure}[t!]
    \centering
    \includegraphics[width=\linewidth]{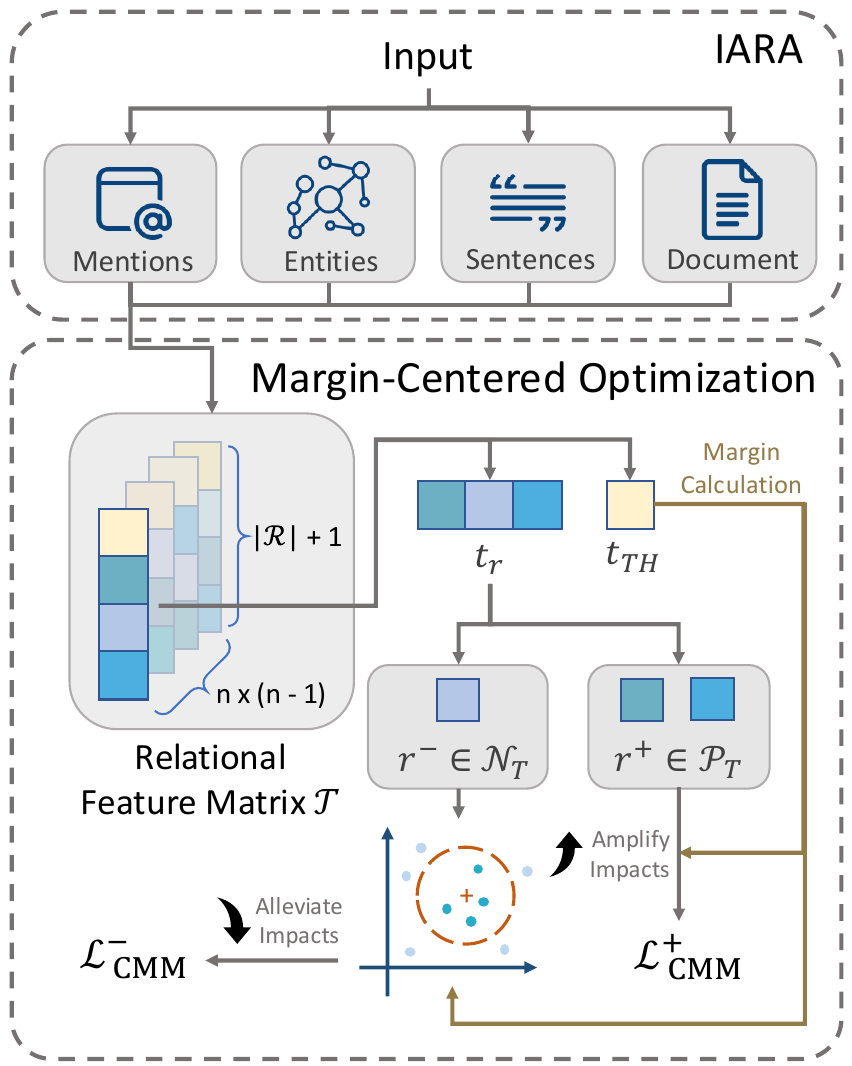}
    \caption{The overall architecture of COMM. 
    Initially, COMM employs the IARA module to capture the relational patterns pertinent to particular entity pairs based on mentions, entities, sentences, and documents. Subsequently, the system transitions to its pivotal phase: optimizing the framework with a focus on margin adjustments. COMM dynamically modifies the margins between relation logits and the decision boundary with the intent to amplify the significance of infrequent positive instances and challenging examples, while concurrently diminishing the influence of the predominant negative instances and straightforward cases.}
    \label{fig:process}
\end{figure}


A significant challenge in DocRE lies in identifying the relevant contextual information for a given entity pair and extracting their relational features. To illustrate this challenge, we present an example from the Re-DocRED dataset \cite{tan-etal-2022-revisiting}, a well-regarded resource in DocRE, as shown in Figure \ref{fig:example}.
This example features a range of complex relations both within and across sentences. 
For instance, determining the \textit{country of citizenship} relation between \textit{Michael Jackson} and \textit{American} requires the DocRE model to focus on a specific sentence and perform intra-sentence relation inference. 
Meanwhile, for inter-sentence relations like the \textit{part of} between \textit{The Way You Make Me Feel} and \textit{Bad}, the model must capture relevant information across multiple sentences to accurately infer the relation types between the entity pair.
To address this challenge, considerable effort is frequently directed toward enhancing sophisticated and advanced relational encoding models \cite{christopoulou-etal-2019-connecting,zeng-etal-2020-double,duan-etal-2022-just,Jain_Mutharaju_Kavuluru_Singh_2024}.
While such efforts have achieved certain success, they often fall short to adequately address the underlying challenge of data. 

In the example shown in Figure \ref{fig:example}, the complete document contains 10 sentences and 23 entities, but only 56 of the entity pairs hold relations, while the remaining 450 entity pairs do not.
To delve deeper into this, we have analyzed the distribution of entity pairs with and without relations (positive samples and negative samples) in the DocRED dataset \cite{yao-etal-2019-docred} and its refined version, Re-DocRED \cite{tan-etal-2022-revisiting}. 
Our findings reveal that in DocRED, only 3.18\% of entity pairs exhibit relations. Even in Re-DocRED, this number only increases to 7.09\%.
The sheer volume of entities and the extensive range of possible relation types pose a significant challenge for human annotators to sustain precision, occasionally resulting in instances where entity pairs that indeed exhibit specific relations are marked as unrelated.
A dataset with a high proportion of negative samples can bias the model toward predicting the majority class, resulting in poor performance in identifying positive relations. 
Moreover, this imbalance extends to the distribution of positive relation categories.
In both DocRED and Re-DocRED, the most frequent positive relation category accounts for over 23\% of samples, while the least frequent category constitutes less than 0.5\%. 
Categories with more samples are typically better optimized, whereas those with fewer samples are often underrepresented, leading to subpar performance for less frequent categories and negatively impacting overall performance.
While some relations are clearly stated, making their accurate inference relatively straightforward, other instances necessitate understanding the context, resolving references, or connecting information dispersed across multiple sentences.
This variation in difficulty adds another layer of complexity to the overall learning process.


To address the diverse challenges presented by the DocRE data and effectively conduct relation inference, we introduce a robust framework called COMM.
COMM begins with an instance-aware reasoning augmentation (IARA) stage, where an existing relational encoding model \cite{zhou2021document, ijcai2021p551, tan-etal-2022-document} is utilized to capture and encode the complex interactions between mentions, entities, sentences, and the document. 
By effectively modeling these interactions, IARA derives contextualized relational feature representations that are finely tuned to the specific instances of entity pairs, ensuring a more precise and instance-aware understanding of the relations within the document.
In the next margin-centered optimization phase, the Concentrated Margin Maximization (CMM) strategy is employed. 
This approach adaptively adjusts the margins between relation logits and the decision boundary to achieve two key objectives: (1) reducing the influence of abundant negative samples while increasing the focus on rare positive samples, and (2) decreasing the emphasis on straightforward instances while boosting the focus on more challenging ones.
This targeted approach directs the model's learning towards more challenging instances and underrepresented classes, thereby enhancing its overall performance.


The key contributions of this work are listed as follows:
\begin{itemize}
    \item To achieve flexibility and efficiency, COMM is formulated as a two-stage framework, featuring instance-aware reasoning augmentation and margin-centered optimization.
    \item To facilitate robust learning, COMM adopts the Concentrated Margin Maximization method to dynamically adjust the margins between relation logits and the decision boundary.
    \item We demonstrate the versatility and effectiveness of COMM through extensive experiments, especially its robustness when trained on low-quality data.
\end{itemize}

\section{Preliminary}

\subsection{Problem Formulation}
In a given document \(D\), with a collection of entities \(\{e_i\}_{i=1}^n\), the objective of DocRE involves determining the types of relations between pairs of entities \((e_s, e_o)_{s,o \in \{1, \ldots, n\}, s \ne o}\), with \(e_s\) representing the subject entity and \(e_o\) the object entity. The possible relations are outlined in the set \(\mathcal{R}\,\cup\,\{\textsc{NA}\}\), where \( \mathcal{R} \) encompasses all predefined relation types and NA is used to indicate the absence of a relation between two entities.

\subsection{Threshold Class}
DocRE can be conceptualized as a multi-label classification problem, wherein each pair of entities within a document may be associated with zero, one, or multiple relations \cite{yao-etal-2019-docred}. 
A straightforward approach to tackle this problem is to employ a battery of binary classifiers, each dedicated to predicting a particular relation type from the set of all possible relations, \(\mathcal{R}\,\). 
These classifiers generate probabilities, \( P(r|e_s, e_o) \), falling within the interval \(\mathcal{I}\), which must be converted into definitive relation labels by applying a threshold mechanism.
Since the determination of the threshold does not have a closed-form solution and is not suitable for differentiation, a common strategy in the field involves sampling various threshold values across the \(\mathcal{I}\) interval and selecting the one that optimizes evaluation metrics such as the \(F_1\) score, a standard metric for relation extraction. 
However, this method may be inadequate due to the model's varying degrees of confidence for different entity pairs or relation types. 
This issue prompted \citet{zhou2021document} to propose the replacement of a fixed global threshold with a learnable, adaptive one. This adaptive threshold is designed to mitigate decision errors during inference by tailoring the threshold to the specific context of each entity pair and relation type.


The relation labels associated with an entity pair \( T = (e_s, e_o) \) can be categorized into positive classes \( \mathcal{P}_T \) and negative classes \( \mathcal{N}_T \). 
The positive classes \( \mathcal{P}_T \) encompass the relations that exist between the entities in \( T \) and are represented as a subset of the total set of relations, \( \mathcal{P}_T \subseteq \mathcal{R} \). In instances where \( T \) does not participate in any relation, \( \mathcal{P}_T \) is defined as an empty set.
Conversely, the negative classes \( \mathcal{N}_T \) are composed of relations that are not exhibited by the entities in \( T \) and are similarly denoted as a subset of the total relations, \( \mathcal{N}_T \subseteq \mathcal{R} \). 
If \( T \) doesn't exhibit any relations, \( \mathcal{N}_T \) encompasses all relations within \( \mathcal{R} \).
\citet{zhou2021document} hypothesize that the accurate classification of an entity pair necessitates that the logits for positive classes exceed a predetermined threshold while those for negative classes remain below this threshold.
Therefore, they introduced an extra class called the threshold class (TH class). 
During training, the logit values for positive relations are gradually optimized to be greater than the logit of the TH class. 
In contrast, the logit values for negative relations need to be smaller than the logit of the TH class. During testing, classes with logits surpassing the TH class are identified as positive classes. In scenarios where such classes are absent, the classification defaults to “NA”.
Due to its verified effectiveness \cite{xie-etal-2022-eider, duan-etal-2022-just, wang-etal-2023-adaptive}, this adaptive threshold class is also incorporated into COMM to detect positive relation instances.

\section{Methodology}
COMM consists of two main stages, namely instance-aware reasoning augmentation (IARA) and margin-centered optimization. The IARA stage facilitates the transition from text to latent relational features, while the margin-centered optimization dynamically refines the entire framework.
The overall architecture of COMM is shown in Figure \ref{fig:process}.

\subsection{Instance-Aware Reasoning Augmentation}
In contrast to sentence-level relation extraction, where typically each sample involves just a single pair of entities, DocRE often deals with a multitude of entities, each of which may be mentioned multiple times throughout the document.
Therefore, the initial step in COMM is to capture and encode the intricate relational patterns between entities within the context of the entire document.
To achieve this goal and enable COMM to efficiently iterate with state-of-the-art DocRE models, we standardize this process and leverage existing DocRE models as relational encoders. 
These models often employ pre-trained language models (PLMs) for contextualized document encoding. 
Subsequently, they use fully connected layers, GCNs, or other techniques to implicitly or explicitly capture interactions across different components at various granularities—such as mentions, entities, sentences, and even the entire document—to comprehensively extract relational features between entities.
However, since addressing DocRE requires using the same number of binary classifiers as the number of relation types, along with an additional threshold class (TH class) that automatically learns whether a relation holds, we need to map the relational features extracted by the relational encoders to a final \(|\mathcal{R}| + 1\) dimension to ensure the next stage proceeds smoothly, where \(|\mathcal{R}|\) denotes the number of distinct relation types.
In other words, at the IARA stage, we use relational encoders along with a transformation layer to map the input document \(D\) and its included \(n*(n-1)\) entity pairs \((e_s, e_o)_{s,o \in \{1, \ldots, n\}, s \ne o}\) into a feature matrix of size \((n*(n-1), |\mathcal{R}| + 1)\).

We can summarize the process as follows:

\begin{equation}\label{eq:maineq1}
    \mathcal{T} = \text{IARA}(\text{REncoder}, D, |\mathcal{R}| + 1)
\end{equation}
where REncoder refers to an advanced relational encoder and \(\mathcal{T}\) is the resultant transformed relation feature matrix of size \((n*(n-1), |\mathcal{R}| + 1)\).

\subsection{Margin-Centered Optimization}
After obtaining the latent relational features \(\mathcal{T}\), we proceed to the second and core stage of COMM which is to dynamically optimize the entire framework based on margin adjustments.
Drawing inspiration from the TH class, COMM seeks to address DocRE by maximizing the distance between the relation logits and the decision boundary.
For clarity, we examine the operation of this stage when presented with a single pair of entities.
We denote the relational feature of the entity pair (\(e_s, e_o\)) as \(\mathcal{T}_{(e_s, e_o)}\) and abbreviate it as \(t\), which is a vector of size \(|\mathcal{R}|+1\). 
In this vector, the first element, at index 0, serves as the logit for the TH class. The subsequent \(|\mathcal{R}|\) elements represent the logits corresponding to the \(|\mathcal{R}|\) specific relation types.
Given our objective to maximize the distance between the relation logits and the decision boundary, the margin loss \cite{rosset2003margin} emerges as a natural and appropriate choice. 
This method encourages a clear distinction between different classes, thereby enhancing the model's generalization capability \cite{deng2019arcface}. 
In our scenario, we strive to maximize the distance between the decision boundary \( t_{\text{TH}} \) and the relation logit \( t_r \), where \( r \in \mathcal{R} \).
The distance between \( t_r \) and \( t_{\text{TH}} \) is defined as:
\begin{equation}\label{eq:dis}
\begin{aligned}
d_{r^+} &= t_{r^+} - t_{\text{TH}}, \quad r^+ \in \mathcal{P}_T \\
d_{r^-} &= t_{\text{TH}} - t_{r^-}, \quad r^- \in \mathcal{N}_T
\end{aligned}
\end{equation}

Building upon the established notations and drawing from the foundational principles of margin loss, the formulation of margin loss in the context of DocRE can be expressed as:
\begin{equation}
\mathcal{L} = \sum_{r \,\in\, \mathcal{P}_T \,\cup \,\mathcal{N}_T} -\,d_r
\label{eq:margin}
\end{equation}

Optimizing this loss is equivalent to maximizing the \(d_r\) variable, which pushes the logit \(t_r^+\) above \(t_{\text{TH}}\) and the logit \(t_r^-\) below \(t_{\text{TH}}\).
However, this straightforward margin loss does not account for the effects of class imbalances or the varying difficulty of instances, potentially leading to suboptimal performance.


\begin{figure}[t]
    \centering
    \includegraphics[width=\linewidth]{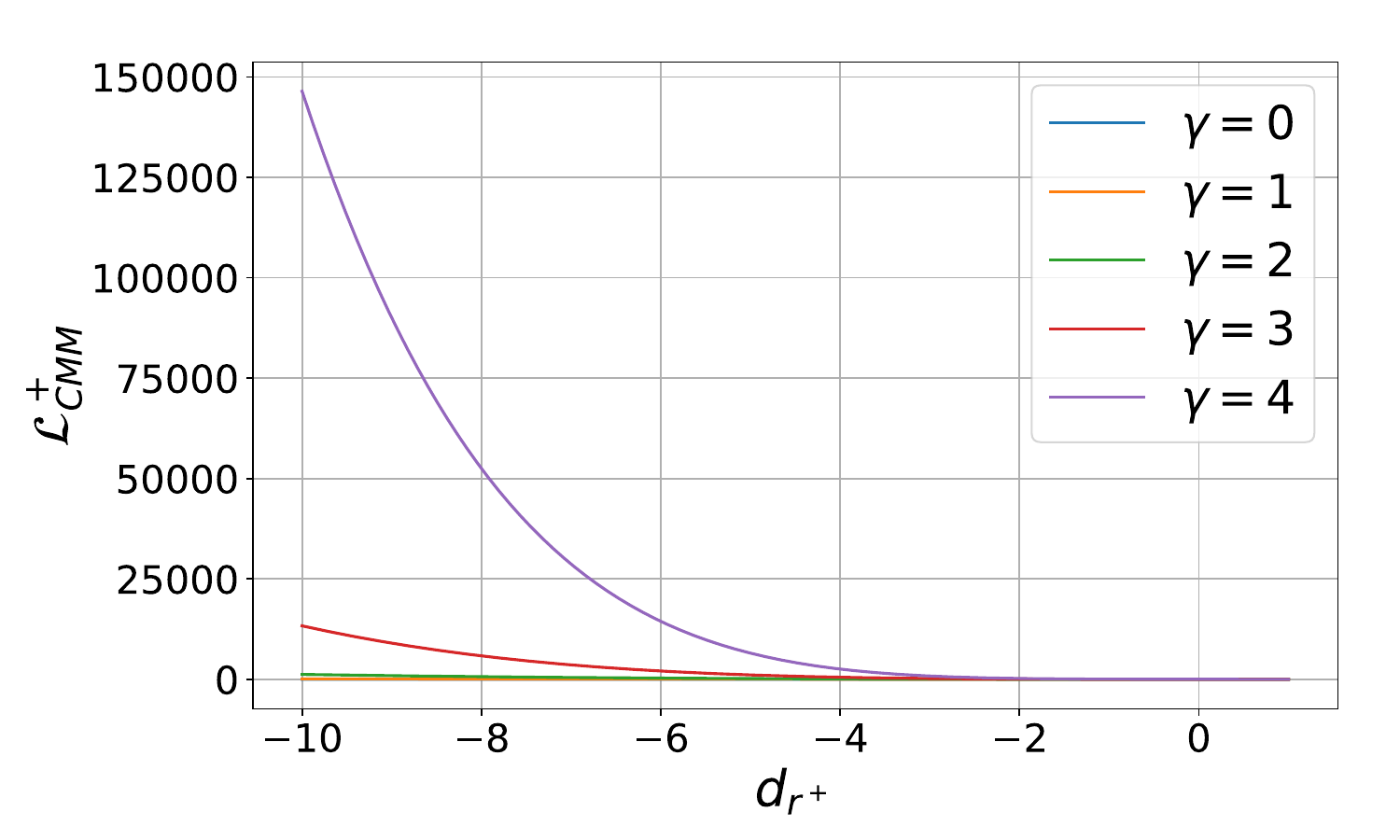}  
    \caption{Visualization of the \(\mathcal{L}_{\,\text{CMM}}\) for positive relations across varying \( \gamma \) values.}
    \label{fig:pos_loss}
\end{figure}

\subsubsection{Concentrated Margin Maximization}
\label{sec:COMMl}

To address the class imbalance issues arising from factors such as mislabeling, and to prevent potential training inadequacies due to the presence of easy and hard instances, we introduce the Concentrated Margin Maximization (CMM) method. 
After computing the distance \(d\) between the relation logits and the adaptive decision boundary using Equation \ref{eq:dis}, CMM rescales this distance as follows:

\begin{equation}\label{eq:rescale}
\begin{aligned}
q_{r^+} &= \log(\sigma(d_{r^+})) \\
q_{r^-} = \log(&\min(\sigma(d_{r^-})+m, 1))
\end{aligned}
\end{equation}
where \(\sigma\) denotes the sigmoid function, and \(m\) is a constant value. 
The range of \(d_r\) calculated using Equation 2 is unknown. 
However, by applying Equation 4, CMM can scale \(d_r\) and convert it to a more defined range, facilitating analysis and enhancing reasoning capabilities.
Subsequently, CMM dynamically increases the weight of positive classes and calculates the overall loss as follows:
\begin{equation}\label{eq:COMMl}
\mathcal{L}_{\,\text{CMM}} = -(\sum_{r^+ \in \mathcal{P}_T}(1 - q_{r^+})^\gamma q_{r^+} + \sum_{r^- \in \mathcal{N}_T} q_{r^-})
\end{equation}
where \(\gamma\) is a hyperparameter.


In general, \(\mathcal{L}_{\,\text{CMM}}\) consists of two primary components, each specifically designed for the positive and negative relation classes, respectively.
As mentioned in the Introduction, the positive relation class tends to comprise a smaller fraction of the training data.
Consequently, it is intuitive to increase the weight of positive examples. 
With the help of the concentration factor \( (1 - q_{r^+})^\gamma \), the positive component of \(\mathcal{L}_{\,\text{CMM}}\) exhibits two notable properties:
(1) In the event of misclassification where the distance metric \( d_{r^+} \) is small, the concentration factor \( (1 - q_{r^+})^\gamma \) takes on a significantly elevated value, thereby increasing the weight of the loss associated with positive samples. On the other hand, when \( d_{r^+} \) is large, signifying a precise prediction, \( (1 - q_{r^+})^\gamma \) converges to 1, leaving the loss unaltered.
(2) The positive component of \(\mathcal{L}_{\,\text{CMM}}\) across various values of \(\gamma\) is visually depicted in Figure \ref{fig:pos_loss}. It is evident from the figure that the parameter \(\gamma\) modulates the rate at which positive examples are up-weighted. 
As \(\gamma\) increases, \(\mathcal{L}_{\,\text{CMM}}\) becomes larger at the same \(d_r^+\), and the gradient becomes steeper. 
Furthermore, irrespective of the \(\gamma\) value, when \(d_r^+\) is considerable, the loss is comparatively low, and the curve exhibits a more gradual incline, suggesting smaller gradients and smaller update steps.
Both aforementioned attributes indicate that the CMM process can not only amplify the influence of positive examples on model optimization but also allocate greater weights to more challenging examples.


We proceed to examine the application of CMM on the classification of negative samples.
In addressing the DocRE task as a multi-label classification problem, the dominance of negative examples over positives in the context of multiple binary classifiers presents a significant hurdle in the learning process for identifying positive relations. 
To address this imbalance, we initially normalize the distance variable \( d_{r^-} \) between \( t_{r^-} \) and \( t_{TH} \) to the range (0, 1) using the sigmoid function.
Following this normalization, we eliminate negative predictions with scores exceeding \( 1-m \) by blocking the accumulation of gradients for these samples. 
After this process, CMM applies the logarithmic function to further rescale it.
The hyperparameter \(m\) controls the cutoff threshold for this filtering process.
This approach is grounded in the rationale of excluding negative instances that are predicted with high confidence and precision, thereby minimizing their overall impact on the learning process.
In addition, when \( d_{r^-} \) is small—indicating that the sample is more difficult and poorly optimized—the value of \(\mathcal{L}_{\,\text{CMM}^-}\) increases, thereby enhancing the impact of these challenging samples on the overall optimization process.


By integrating these two components, we obtain the complete dynamic adjustment process of CMM and derive the full definition of \(\mathcal{L}_{\,\text{CMM}}\), as detailed in Equation \ref{eq:COMMl}.
This formulation is intended to enhance the learning mechanism by emphasizing both positive and challenging instances while reducing the influence of confidently predicted positives and negatives.
As a result, it fosters a more balanced optimization process within COMM.

\begin{algorithm}[t]
\caption{COMM}
\textbf{Input}: REncoder, BatchData, Relations \(\mathcal{R}\) \\
\textbf{Parameter}: $\gamma$, $m$ 
\label{algo:workflow}
\begin{algorithmic}[1]
\FORALL {\(D\) \textbf{in} BatchData}
    \STATE $\mathcal{L}_{\,\text{CMM}} \gets 0$
    \STATE $\mathcal{T} \gets \text{IARA(\text{REncoder}, D, \(|\mathcal{R}| + 1\))}$
    \FORALL {Entity Pair $(e_s, e_o)$ \textbf{in} \(D\)}
        \STATE $t_{TH} \gets \mathcal{T}(e_s, e_o)[0]$
        \FORALL {$r$ \textbf{in} \(\mathcal{R}\)}
            \IF {$r \in \mathcal{P}_T$}
                \STATE $t_{r^+} \gets \mathcal{T}(e_s, e_o)[r]$
                \STATE $d_{r^+} \gets t_{r^+} - t_{TH}$
                \STATE $q_r^+ \gets \log(\sigma(d_{r^+}))$
                \STATE $\mathcal{L}_{\,\text{CMM}} \gets \mathcal{L}_{\,\text{CMM}} - q_{r^+}(1 - q_{r^+})^{\gamma}$
            \ELSIF {$r \in \mathcal{N}_T$}
                \STATE $t_{r^-} \gets \mathcal{T}(e_s, e_o)[r]$
                \STATE $d_{r^-} \gets t_{TH} - t_{r^-}$
                \STATE $q_{r^-} \gets \log(\min(\sigma(d_{r^-}) + m, 1))$
                \STATE $\mathcal{L}_{\,\text{CMM}} \gets \mathcal{L}_{\,\text{CMM}} - q_{r^-}$
            \ENDIF
        \ENDFOR
    \ENDFOR
    \STATE \textbf{Perform Stochastic Gradient Descent using} $\mathcal{L}_{\,\text{CMM}}$
\ENDFOR
\end{algorithmic}
\end{algorithm}

\subsection{Workflow}
The workflow of the COMM framework is illustrated in Algorithm \ref{algo:workflow}. 
Initially, we employ IARA within COMM to encode the document and entity pairs, thereby obtaining the latent relational features for each entity pair. 
Subsequently, we utilize the Concentrated Margin Maximization approach to dynamically adjust the margins and optimize the model via \(\mathcal{L}_{\,\text{CMM}}\). 
This optimization process increases the weight of positive instances, reduces the influence of abundant negative examples, and enhances learning from challenging samples.
Finally, during the prediction phase, we employ the dynamic logit value \( t_{TH} \) of the TH class to predict the relation categories. 
Specifically, a relation is considered present if its logit \( t_{r} \) exceeds \( t_{TH} \). If no such relation exists, the model concludes that there is no relation between the entities and returns NA.

\begin{table*}[t!]
    \centering
    \setlength{\tabcolsep}{1mm}
    \begin{tabular}{lcclccclc}
        \toprule
        \multicolumn{1}{l}{\multirow{2}{*}{Method}} & \multicolumn{4}{c}{DocRED-Mixed} & \multicolumn{4}{c}{Re-DocRED} \\
        \multicolumn{1}{c}{} & \makecell{dev $F_1$} & \makecell{dev $F_1$ Ign} & test $F_1$ & \makecell{test $F_1$ Ign} & \makecell{dev $F_1$} & \makecell{dev $F_1$ Ign} & test $F_1$ & \makecell{test $F_1$ Ign} \\
        \midrule
        \multicolumn{1}{l}{BERT-ATLOP} & 50.5 & 49.9 & 50.1 & 49.5 & 73.3 & 72.6 & 72.9 & 72.2 \\
        \rowcolor{mygray}
        \multicolumn{1}{l}{COMM-BERT-ATLOP} & {61.8} & {60.7} & {62.3}{(+12.2)} & {61.3} & {76.3} & {75.0} & {76.1}{(+3.2)} & {74.8} \\
        \midrule
        \multicolumn{1}{l}{RoBERTa-ATLOP} & 51.8 & 51.3 & 50.6 & 50.2 & 77.6 & 77.0 & 77.5 & 76.9 \\
        \rowcolor{mygray}
        \multicolumn{1}{l}{COMM-RoBERTa-ATLOP} & {63.9} & {62.7} & {63.5}{(+12.9)} & {62.4} & {80.8} & {79.7} & {80.8}{(+3.3)} & 79.7 \\
        \midrule
        \multicolumn{1}{l}{BERT-DocuNet} & 47.9 & 47.5 & 46.9 & 46.5 & 75.3 & 74.4 & 75.0 & 74.1 \\
        \rowcolor{mygray}
        \multicolumn{1}{l}{COMM-BERT-DocuNet} & {60.9} & {59.8} & {60.6}{(+13.7)} & {59.5} & {76.4} & {75.0} & {76.3}{(+1.3)} & 75.0 \\
        \midrule
        \multicolumn{1}{l}{RoBERTa-DocuNet} & 48.6 & 48.4 & 48.7 & 48.5 & 78.1 & 77.3 & 78.5 & 77.9 \\
        \rowcolor{mygray}
        \multicolumn{1}{l}{COMM-RoBERTa-DocuNet} & {62.9} & {61.9} & {62.1}{(+13.4)} & {61.1} & {79.4} & {78.4} & {79.5}{(+1.0)} & 78.4 \\
        \midrule
        \multicolumn{1}{l}{BERT-KD-DocRE} & 52.2 & 51.7 & 51.3 & 50.9 & 75.2 & 74.3 & 75.0 & 74.1 \\
        \rowcolor{mygray}
        \multicolumn{1}{l}{COMM-BERT-KD-DocRE} & {60.9} & {59.4} & {60.8}{(+9.5)} & {59.5} & {75.9} & {74.3} & {76.1}{(+1.1)} & {74.6} \\
        \midrule
        \multicolumn{1}{l}{RoBERTa-KD-DocRE} & 54.6 & 54.2 & 54.9 & 54.5 & 78.7 & 77.8 & 78.5 & 77.6 \\
        \rowcolor{mygray}
        \multicolumn{1}{l}{COMM-RoBERTa-KD-DocRE} & {63.3} & {62.3} & {63.1}{(+8.2)} & {62.2} & {79.9} & {78.8} & {79.5}{(+1.0)} & {78.5} \\
        \bottomrule
    \end{tabular}
    \caption{Evaluation of COMM's efficacy when ATLOP \cite{zhou2021document}, DocuNet \cite{ijcai2021p551}, and KD-DocRE \cite{tan-etal-2022-document} are used as relational encoders within the IARA module, respectively.
    All models are trained utilizing the training data of either DocRED or Re-DocRED. For evaluation purpose, we utilize the Re-DocRED dev and test sets, which offer superior annotation quality and can provide a more accurate reflection of the model's capabilities. All the results are obtained by averaging three repeated executions.
}
    \label{tab:mainres}
\end{table*}

\section{Experiments}

\subsection{Datasets}
To fully evaluate COMM, we conduct comprehensive experiments based on two widely adopted DocRE datasets. 
\begin{itemize}
    \item DocRED \cite{yao-etal-2019-docred} represents a comprehensive, crowd-sourced dataset for DocRE created from Wikipedia articles. Comprising 3,053 documents for training purposes, it spans a diverse array of domains. This dataset demands that DocRE models exhibit a wide range of reasoning capabilities, including but not limited to coreference resolution and common-sense reasoning.
    \item Re-DocRED \cite{tan-etal-2022-revisiting} constitutes an augmented variant of the DocRED dataset, specifically designed to rectify the prevalent issue of false negatives in DocRE. Through the process of re-annotation of the original dataset, Re-DocRED substantially amplifies the quantity and scope of annotated relation triples. This enhancement greatly contributes to the overall improvement in the dataset's quality and comprehensiveness.
\end{itemize}

\subsection{Implementation Details}
COMM is implemented utilizing the PyTorch framework \cite{paszke2019pytorch}. 
We have incorporated several publicly accessible and influential DocRE models as the relational encoder in the IARA stage of COMM, including ATLOP \cite{zhou2021document}, DocuNet \cite{ijcai2021p551}, and KD-DocRE \cite{tan-etal-2022-document}. 
All results are reported based on the public code of these models with three repeated executions.
The parameter \(m\) is selected through a grid search over the values [0.1, 0.2, 0.3, 0.4], while \(\gamma\) is chosen through [1, 1.2, 1.4, 1.6, 2]. The COMM framework is optimized using the AdamW optimizer \cite{loshchilov2017decoupled}.

\subsection{Evaluation Setup}
For training, we utilize the training sets of both DocRED and Re-DocRED, each comprising 3,053 documents. During validation and testing phases, in order to better assess the model's performance, we employ the higher-quality Re-DocRED validation and test sets, each consisting of 500 documents.
In subsequent discussions, these distinct configurations will be referred to as \textbf{DocRED-Mixed} and \textbf{Re-DocRED}, respectively.
The reason for this setup is that Re-DocRED, as an improved version of DocRED, includes a substantial amount of annotations that were originally false negatives. 
This enhancement allows for a more comprehensive evaluation of the model's performance using such data.
In line with \citet{yao-etal-2019-docred}, we evaluate the performance using both $F_1$ score and Ign $F_1$ score. The $F_1$ Ign score represents the $F_1$ metric calculated after excluding those relational facts that are common to both the training set and the dev/test sets.

\subsection{Results of COMM}
The test results for COMM and several baseline methods under the DocRED-Mixed and Re-DocRED settings are summarized in Table \ref{tab:mainres}. It is evident that when trained with the higher-quality Re-DocRED data, both COMM and various baseline models can maintain a relatively high \(F_1\) score, exceeding 70\%. 
Despite this, COMM demonstrates notable enhancements across the board, with performance gains ranging from 1.0\% up to a significant 3.3\%.
Such enhancements are particularly striking, as they have been achieved without the extensive architectural modifications that have traditionally been necessary for progress in this domain \cite{duan-etal-2022-just,ma-etal-2023-dreeam,zhu-etal-2024-fcds-fusing,Jain_Mutharaju_Kavuluru_Singh_2024}. 
Instead, COMM stands out as a streamlined framework, primarily targeting enhancements in the optimization process.
When using the lower-quality DocRED dataset, where only 3.18\% of entity pairs in the training data are associated with relations, the performance of baseline methods declines sharply, dropping from over 70\% to around 50\%, highlighting the challenging nature of the DocRED-Mixed setting. 
In comparison, employing COMM significantly enhances performance in the DocRED-Mixed setting, yielding substantial improvements ranging from 8.2\% to 13.7\% over the baseline approaches.
The outcomes underscore the robustness and efficacy of the COMM framework, demonstrating its potential as a crucial tool in addressing the challenges of imbalanced data in DocRE.

\begin{table}[t]
    \centering
    
    \begin{tabular}{lcc}
  	\toprule
  	\multicolumn{1}{l}{Loss function} & \makecell{$F_1$} & \makecell{$F_1\,{\text{Ign}}$}\\
  	\midrule
        ATL~\cite{zhou2021document} & 50.09 & 49.50 \\
        Balanced Softmax~\cite{ijcai2021p551}& 46.25 & 45.83 \\
        AML~\cite{wei2022sagdre} & 43.21 & 43.04 \\
        AFL~\cite{tan-etal-2022-document} & 50.58 & 50.23 \\
        HingeABL~\cite{wang-etal-2023-adaptive} & 58.44 & 57.45 \\
        \midrule
        CMM Loss & \textbf{62.31} & \textbf{61.33} \\
    \bottomrule
  \end{tabular}
  \caption{Comparison of CMM loss with other widely adopted loss functions. The results are obtained based on COMM-BERT-ATLOP$_{\text{BASE}}$ under the DocRED-Mixed setting by substituting different loss functions.
}
    \label{tab:t2}
\end{table}

\subsection{Comparison of Loss Functions}

In our preceding experiments and analyses, we have substantiated the robustness and efficacy of our COMM framework.
As the core component of COMM, the impact of the Concentrated Margin Maximization strategy is primarily reflected in the dynamic scaling of the loss function.
To thoroughly evaluate its design, we have conducted a comparative analysis of the CMM loss against various commonly employed loss functions in the DocRE domain \cite{zhou2021document, ijcai2021p551, wei2022sagdre,tan-etal-2022-document, wang-etal-2023-adaptive} based on COMM-BERT-ATLOP$_{\text{BASE}}$.
The results shown in Tables \ref{tab:t2} and \ref{tab:t3} indicate that CMM loss outperforms other loss functions, achieving a gain of over 3.87 \(F_1\) in the DocRED-Mixed setting and 0.97 \(F_1\) in the Re-DocRED setting.
Due to different optimization goals, the evaluated loss functions exhibit diverse optimization strategies. 
Specificly, CMM loss is designed based on certain observations of DocRE data to address the imbalance problems.
These results suggest that there is room for improvement in the utilization and exploration of DocRE data.
By dynamically adjusting the weights of samples, the impact of data biases can be mitigated, leading to more robust model training.


\subsection{Effect of CMM Loss on Relation Distribution}

\begin{figure}[t]
    \centering
    \includegraphics[width=\linewidth]{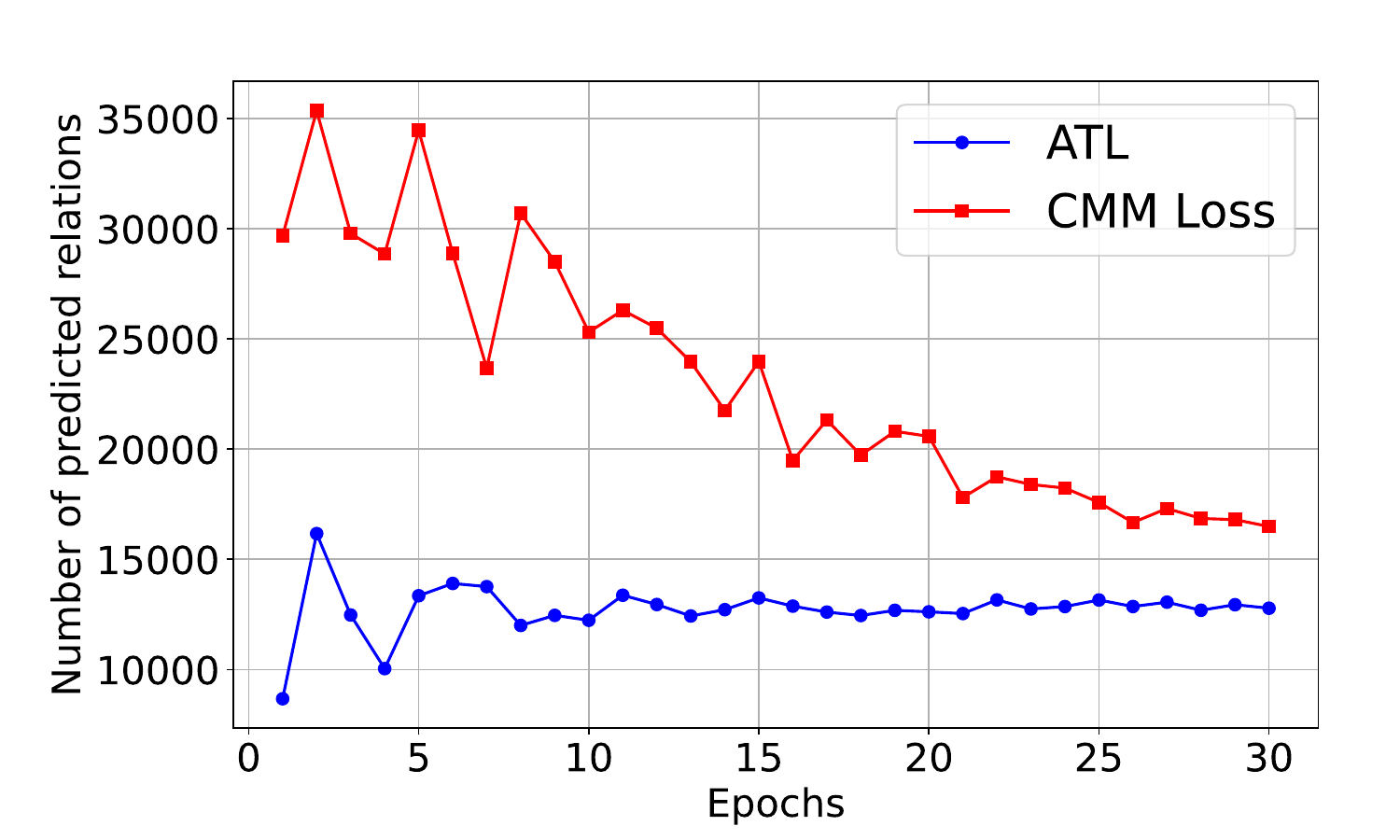}  
    \caption{Number of predicted positive relations versus training epochs. The results are obtained based on COMM-BERT-ATLOP$_{\text{BASE}}$ under the DocRED-Mixed setting.}
    \label{fig:numrels}
\end{figure}

A primary motivation behind the development of CMM is to address the issue of class imbalance prevalent in DocRE. 
Although the experiments in the previous section have already demonstrated the effectiveness of the CMM loss compared to other loss functions, it is still uncertain whether it adjusts the model in the direction we anticipated.
If the CMM loss indeed improves model performance by increasing the weight of positive samples, the number of positive instances predicted should significantly increase. 
Under the DocRED-Mixed setting, we optimize COMM-BERT-ATLOP$_{\text{BASE}}$ using either ATL \cite{zhou2021document} (a significant and widely adopted loss function in the field) or CMM loss separately and track the number of positive relation predictions made by each approach as training proceeds.
The outcomes are illustrated in Figure \ref{fig:numrels}.
The visualization clearly demonstrates that as training progresses, the model optimized with CMM loss predicts a significantly greater number of positive relations compared to the one optimized with ATL. 
The result proves that CMM can indeed increase the weight of positive samples, encouraging the model to generate more positive predictions, thereby balancing the impact of data imbalance.
The trend in the number of positive relations predicted by the two methods suggests that while the model trained with ATL shows less change in the count of positive relations, the model trained with CMM loss is able to produce a significant number of positive predictions from the initial stages and continues to refine its predictions over time. 
This indicates that CMM loss is not only effective in increasing the recall of positive relations but also supports the model's ability to discriminatively learn from the training data, thereby enhancing its overall performance in identifying relational information within documents.
These observations, coupled with the performance gap between ATL and CMM loss, as illustrated in Tables 2 and 3, underscores the critical importance of effectively balancing the learning of positive and negative classes in the context of DocRE.

\begin{table}[t]
    \centering
    
    \begin{tabular}{lcc}
  	\toprule
  	\multicolumn{1}{l}{Loss function} & \makecell{$F_1$} & \makecell{$F_1\,{\text{Ign}}$}\\
  	\midrule
        ATL~\cite{zhou2021document} & 72.87 & 72.22 \\
        Balanced Softmax~\cite{ijcai2021p551}& 73.57 & 72.79 \\
        AML~\cite{wei2022sagdre} & 72.96 & 72.14 \\
        AFL~\cite{tan-etal-2022-document} & 73.71 & 72.87 \\
        HingeABL~\cite{wang-etal-2023-adaptive} & 75.15 & 73.84 \\
        \midrule
        CMM Loss & \textbf{76.12} & \textbf{74.76} \\
    \bottomrule
  \end{tabular}
  \caption{Comparison of CMM loss with other widely adopted loss functions. The results are obtained based on COMM-BERT-ATLOP$_{\text{BASE}}$ under the Re-DocRED setting by substituting different loss functions.}
    \label{tab:t3}
\end{table}

\section{Related Work}

Relation extraction is a crucial component in many real-world applications \cite{hixon-etal-2015-learning,hong2020novel, duan2021bridging, li2023toward, li2024flexkbqa, li2024focusllm}. As noted by \citet{yao-etal-2019-docred}, a great proportion of relation instances can only be deduced from the multi-sentence context. DocRE, which presents a more realistic setting, has recently garnered increasing attention.
To address the challenges posed by DocRE, such as intricate context, the majority of existing research tends to focus on model architectures \cite{zhou2021document, duan-etal-2022-just, xie-etal-2022-eider, tan-etal-2022-document, ma-etal-2023-dreeam, zhu-etal-2024-fcds-fusing}. 
A pioneering study in this field is conducted by \citet{zhou2021document}. This work not only successfully extracts entity and entity pair features based on attention scores from pretrained language models (PLMs) \cite{devlin2018bert, liu2019roberta} but also introduces the novel concept of adaptive threshold class. 
This advancement negates the requirement for threshold searching via dev sets, thereby establishing a foundational framework that has guided subsequent research in the area.
Efforts to refine model structures to better solve DocRE can be broadly categorized into two types.
The first category either implicitly inherits the attention distribution learned within PLMs or designs additional attention modules to capture long-range dependencies \cite{ye-etal-2020-coreferential,zhou2021document,ijcai2021p551,yuan2021document,xu2021entity,yu-etal-2022-relation}.
Models in this category, primarily structured around PLMs, are relatively simple in architecture and require less computational expense for training. 
Despite these advantages, their performance is generally less notable due to an inadequate representation and modeling of the relations among entities and sentences.
The second category, on the other hand, usually employs Graph Convolutional Networks (GCNs) \cite{kipf2016semi} to capture complex interactions among various components (e.g., mentions, entities, sentences) and to facilitate reasoning \cite{yang2021sagcn,zeng-etal-2021-sire, duan-etal-2022-just,wei2022sagdre,liu2023document,zhu-etal-2024-fcds-fusing}. 
Models in this category typically achieve superior results, although the incorporation of GCNs makes them more computationally intensive and complex in architecture.
Beyond focusing on model architectures, there are also studies that aim to enhance the performance of DocRE from different perspectives, such as loss function \cite{wang-etal-2023-adaptive} and data augmentation \cite{tan-etal-2022-revisiting}.

\section{Conclusions}

In this paper, we have introduced COMM, a flexible framework for DocRE.
COMM consists of two main components, namely instance-aware reasoning augmentation (IARA) and margin-centered optimization. 
With the support of existing advanced DocRE models, IARA extracts contextualized features from documents and maps them to a relational feature matrix.
Following this, in the margin-centered optimization stage, COMM employs the Concentrated Margin Maximization (CMM) method to enhance the weights of positive samples, thereby addressing the impact of class imbalance between positive and negative samples. 
Additionally, CMM takes into account the varying difficulty levels of instances and dynamically adjusts their influence on model training. 
Our experimental results underscore the robustness of COMM, showcasing its substantial performance improvements, particularly in handling low-quality data, with enhancements exceeding 10\%.

\section*{Acknowledgments}
This work was supported in part by National Key Research and Development Program of China under Grant No. 2020YFA0804503, National Natural Science Foundation of China under Grant No. 62272264, and Beijing Academy of Artificial Intelligence (BAAI).

\bibliography{aaai25}

\end{document}